\documentclass[preprint,12pt]{elsarticle}
\journal{Medical Image Analysis}

\usepackage[T1]{fontenc}
\usepackage{graphicx}
\usepackage{amsmath}
\usepackage{amssymb}
\usepackage[ruled, vlined]{algorithm2e}
\usepackage{algpseudocode}
\usepackage{booktabs}
\usepackage{pifont}
\newcommand{\xmark}{\ding{55}}%
\usepackage{subcaption}
\usepackage{lineno}
\usepackage[table]{xcolor}   %
\usepackage{geometry}

\newcommand{\revision}[1]{\textcolor{black}{#1}}

\usepackage{siunitx}          %

\usepackage{hyperref}
\usepackage{xcolor}

\hypersetup{
    colorlinks,
    linkcolor={blue!50!black},
    citecolor={blue!50!black},
    urlcolor={blue!80!black}
}

\begin{document}
\begin{frontmatter}

\title{Individualized  Mapping of Aberrant Cortical Thickness via Stochastic Cortical Self-Reconstruction}

\author[label1,label2]{Christian Wachinger\corref{cor1}}
\cortext[cor1]{Corresponding author}
\ead{christian.wachinger@tum.de}
\author[label4]{Dennis M. Hedderich}
\ead{dennis.hedderich@tum.de}
\author[label4]{Melissa Thalhammer}
\ead{melissa.thalhammer@tum.de}
\author[label1,label2]{Fabian Bongratz}
\ead{fabi.bongratz@tum.de}
\affiliation[label1]{organization={Lab for Artificial Intelligence in Medical Imaging, Institute for Diagnostic and Interventional Radiology, School of Medicine and Health, TUM Klinikum, Technical University of Munich (TUM)},
            city={Munich},
            postcode={81675},
            country={Germany}}

\affiliation[label2]{organization={Munich Center for Machine Learning (MCML)},
            city={Munich},
            country={Germany}}
\affiliation[label4]{organization={Department of Neuroradiology, School of Medicine and Health, TUM Klinikum, Technical University of Munich (TUM)},
            city={Munich},
            postcode={81675},
            country={Germany}}
\begin{abstract}
    Understanding individual differences in cortical structure is key to advancing diagnostics in neurology and psychiatry. 
Reference models aid in detecting aberrant cortical thickness, yet site-specific biases limit their direct application to unseen data, and region-wise averages prevent the detection of localized cortical changes. 
To address these limitations, we developed the Stochastic Cortical Self-Reconstruction (SCSR), a novel method that leverages deep learning to reconstruct cortical thickness maps at the vertex level without needing additional subject information. Trained on over 25,000 healthy individuals, SCSR generates highly individualized cortical reconstructions that can detect subtle thickness deviations. 
Our evaluations on independent test sets demonstrated that SCSR achieved significantly lower reconstruction errors and identified atrophy patterns that enabled better disease discrimination than established methods. 
It also hints at cortical thinning in preterm infants that went undetected by existing models, showcasing its versatility. 
Finally, SCSR excelled in mapping highly resolved cortical deviations of dementia patients from clinical data, highlighting its potential for supporting diagnosis in clinical practice. 
\end{abstract}


\end{frontmatter}

\section{Introduction}

Understanding individual differences in brain structure, particularly within the cerebral cortex, is essential for advancing diagnostic tools in clinical neurology and psychiatry \cite{frisoni2010clinical}. The cerebral cortex plays a central role in neurodegenerative, psychiatric, and developmental disorders, where cortical abnormalities are often prominent~\cite{matsumoto2023cerebral,van2018cortical,marek2022reproducible,hibar2018cortical,volpe2009brain}. With mental illness and dementia representing the largest global health burden~\cite{vigo2016estimating,arias2022quantifying}, the ability to precisely assess individual brain variations is critical for early diagnosis and effective treatment. However, significant variability in cortical features across individuals, compounded by measurement inconsistencies, poses a critical challenge in using neuroimaging metrics as direct biomarkers \cite{potvin2017normative,frangou2022cortical,alfaro2021confound,wachinger2021detect}.

Brain charts \cite{bethlehem2022brain} and normative brain models \cite{marquand2019conceptualizing,rutherford2022normative,bethlehem2020normative} have been developed to define normative ranges of cortical thickness. They present a standardized approach to detect deviations from typical brain structure by computing centiles or Z-scores. The versatility of these reference models is evident in their application across a range of neurological and psychiatric disorders, including dementia, depression, psychosis, schizophrenia, bipolar disorder, autism, and ADHD \cite{rutherford2023evidence,marquand2019conceptualizing,bethlehem2020normative,ziegler2014individualized,potvin2017normative}. 
\revision{The clinical potential of MRI volumetry for anomaly detection is further demonstrated by the introduction of commercial FDA-cleared volumetry tools.}
Unlike traditional case-control studies, which often assume distinct and homogeneous clinical groups, reference models account for the heterogeneity within patient populations and the shared transdiagnostic pathogenic mechanisms, particularly in psychiatric conditions \cite{verdi2021beyond,opel2020cross}. This offers a more nuanced and comprehensive understanding of brain disorders.

Despite these advances, current reference models still fall short of providing the quantitative precision necessary for accurate diagnosis of individual magnetic resonance scans in clinical practice \cite{bethlehem2022brain}.
This is rooted in a fundamental limitation of these models in explicitly modeling the relationship between covariates and brain measurements \cite{rutherford2022normative}, commonly: $(\text{age, sex, site}) \mapsto \text{cortical phenotype}$. 
Firstly, they cannot directly be used to analyze scans from new imaging sites without adapting to site-specific variations, requiring a collection of over 100 subjects from a new site \cite{bethlehem2022brain}. This is due to the complexity of modeling scanning and processing biases explicitly and instead encapsulating them in a \emph{site} variable \cite{alfaro2021confound,wachinger2021detect}. Secondly, basing reference comparisons solely on factors like age and sex offers a relatively coarse characterization of an individual \cite{foulkes2018studying}. Ideally, we would compare to a hypothetical healthy counterpart, or counterfactual \cite{pearl2009causality}, reflecting what a patient's cortex would look like without disease. This issue becomes even more pronounced when referencing infants, as age-based comparisons may be inaccurate, particularly for those born preterm \cite{villar2018monitoring}.

Additionally, the limited spatial resolution of current reference models prevents the detection of localized cortical abnormalities. The BrainChart \cite{bethlehem2022brain} focuses on overall cortical thickness, and other approaches perform region-based analyses derived from brain atlases \cite{rutherford2022charting}. 
\revision{
However, the validity of these atlas-defined regions has been questioned, particularly because the use of different atlases, each with distinct parcellation schemes, can lead to inconsistent results. Furthermore, they can obscure focal pathology by aggregating vertex-wise data into coarse regional summaries, potentially leading to a loss of important spatial information~\cite{furtjes2023quantified}.}
While it is theoretically possible to extend current reference models to the vertex level, this would require estimating a massive number of independent regression models. 
\revision{Importantly, it is well-established that strong correlations exist across the cortex \cite{friston2011functional} -- not only in adjacent parts but also across more distant areas as captured by structural covariance networks \cite{alexander2013imaging,evans2013networks} -- making the current model framework insufficient to capture these spatial relationships.}

\begin{figure}[th!]
    \centering
    \includegraphics[width=\textwidth]{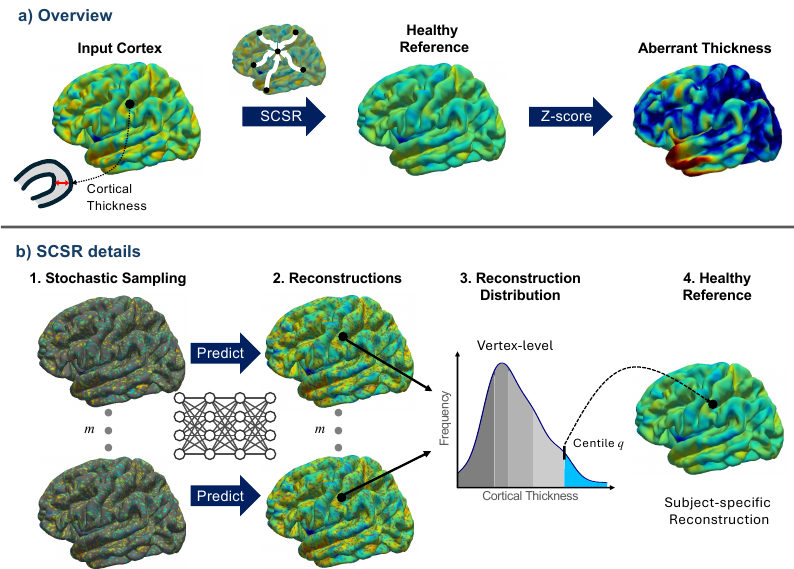}
    \caption{(a) The cortical surface mesh with thickness values extracted from the MRI scan serves as input. 
    The healthy cortical reference is estimated from the input mesh with the stochastic cortical self-reconstruction (SCSR) based on learned spatial relationships (white arrows). The vertex-wise Z-score between input and reference thickness yields a deviation map indicating aberrant thickness. 
    (b) SCSR stochastically samples vertices (colored vertices on the left surface) and predicts the thickness values of the remaining vertices (colored vertices on the right surface) with a deep neural network.  
    The stochastic sampling and prediction are repeated $m$ times to create a distribution over healthy reconstructions. 
    The shading illustrates the 25th, 50th, 75th, and 95th centiles of the thickness distribution for one vertex (black dot). 
    The subject-specific healthy reference is obtained by selecting thickness values from the distribution at a given centile $q$. %
    Note that the only input at inference is the cortical thickness map of the individual. 
    }
    \label{fig:overview}
\end{figure}

We developed a new concept for cortical reference modeling that needs no subject-specific information aside from the \emph{cortex itself} and identifies highly resolved deviations at the vertex level. Our approach leverages deep neural networks to self-reconstruct cortical thickness from a stochastic sampling pattern, called Stochastic Cortical Self-Reconstruction (SCSR), see Fig.~\ref{fig:overview}. By learning the inherent spatial correlations across the cortex, the network can accurately infer missing information and reconstruct the full cortex, even though only a fraction of the cortex is used as input. Like traditional normative models, SCSR is trained on a large dataset of healthy individuals, ensuring that the reconstructed cortex represents a healthy reference. Once trained, SCSR repeatedly reconstructs the cortex of a test scan with varied stochastic sampling patterns. This process generates a distribution of healthy reconstructions tailored to the individual. 
We obtain a single cortical reconstruction by selecting thickness values at the desired centile of the reconstruction distribution. 
Comparing the reconstruction to the actual thickness map results in deviation scores, offering a highly individualized and detailed assessment of the cortex. 

In our experiments, we trained SCSR using data from the large, population-based UK Biobank (UKB) imaging study and demonstrated its ability to identify pathological atrophy patterns in four independent datasets of patients with Alzheimer's disease (AD). 
Quantitative results confirmed the significantly lower reconstruction errors of SCSR, the significant decrease in Z-scores as dementia progresses, and improved disease discrimination. 
Additionally, we applied SCSR to an in-house clinical dataset containing four types of dementia: AD, posterior cortical atrophy (PCA), behavioral variant frontotemporal dementia (bvFTD), and semantic dementia (SD). The results showed that dementia-specific atrophy patterns clearly emerged from individual cortical deviation maps at an individual level. 
Beyond adults, we trained a region-based variant of SCSR on infants and detected significant regional cortical thinning in preterm-born infants, which was not suggested by existing reference models. 

All adult experiments were conducted in an \emph{out-of-domain} setting, where the evaluation dataset differed from the training dataset. The high performance of SCSR in these challenging conditions can be attributed to SCSR operating on thickness values, such that all scanning- and processing-related nuisance variables, which typically confound the relationship between age, sex, and cortical measurements, are already present in the input. Hence, we bypass the need for explicitly modeling the myriad potential sources of bias \cite{alfaro2021confound}. Rather, the intricacies of an individual's cortex offer the most pertinent insights into the expected healthy reference. By leveraging the complex relationship of cortical structure, SCSR offers a promising approach for supporting clinical diagnosis. %

\section{Methods}

\subsection{Data \label{sec:data}}
\textbf{Adults:} We used several large-scale public neuroimaging datasets and an in-house dataset for our experiments on adults, encompassing subjects from four continents and various types of dementia. 
For training the adult model, we used healthy subjects from the UK Biobank Imaging study \cite{littlejohns2020uk}, see Table~\ref{tab:datasets} for subject characteristics and Sec.~\ref{sec:datadetails} for selection procedure. 
As part of the evaluation, we used four datasets with AD patients: the Alzheimer’s Disease Neuroimaging Initiative (ADNI) \cite{jack2008alzheimer}, the Japanese ADNI (J-ADNI) \cite{iwatsubo2010japanese},  the Australian Imaging, Biomarkers and Lifestyle (AIBL) study \cite{ellis2009australian}, and the 
DZNE-Longitudinal Cognitive Impairment and Dementia Study (DELCODE) \cite{jessen2018design}, with details in Sec. \ref{sec:datadetails}. 
In addition, we used routine clinical data from 50 subjects,  with an equal distribution of 10 patients per group across AD, posterior cortical atrophy (PCA), behavioral variant frontotemporal dementia (bvFTD), semantic dementia (SD), and normal cognition (CN) for evaluating differential diagnosis. 
These patients were selected from our hospital database, and diagnosis was established using clinical criteria, FDG-PET metabolism information, and cerebrospinal fluid biomarkers.
All scans (T1w MRI) were processed with FreeSurfer \cite{fischl2012freesurfer} to estimate cortical thickness. 
For working on the vertex level, spatial cortex correspondences across subjects need to be established. 
We follow the procedure in FreeSurfer with inflation (\texttt{mris\_inflate}), mapping to the sphere (\texttt{mris\_sphere}), spherical registration (\texttt{mris\_register}), and resampling  (\texttt{mri\_surf2surf}) (see Fig. \ref{fig:freesurfer}) \cite{fischl1999cortical}. The registration is performed to the FsAverage template~\cite{fischl1999fsaverage}, and the target for resampling is an icosphere of order five, yielding 10,242 cortical thickness measures per hemisphere. %
Automated QC was performed based on the Euler characteristic~\cite{fischl2001automated}, which represents the number of topological defects and is consistently correlated with visual quality control ratings \cite{rosen2018quantitative,monereo2021quality}. %
Outliers are identified based on the interquartile range method on the Euler characteristics similar to \cite{backhausen2022best,zugman2022mega,rutherford2022charting}.
We randomly split the UKB data into training (80\%; 25,338 subjects) and validation (20\%; 6,334 subjects). 
For the ADNI dataset, we randomly selected 20\% of control and AD subjects for validation, and the rest was used as a test set. 
The in-house, AIBL, J-ADNI, and DELCODE datasets were entirely used as indepdendent test sets. %

\noindent
\textbf{Infants:} Infant brain measures were obtained from the developing Human Connectome Project (dHCP) \cite{hughes2017dedicated} that collected data from fetal and/or neonatal participants. 
We used neuroimaging data from 375 term-born and 92 preterm-born infants, who were scanned at term-equivalent age (37 – 45 weeks postmenstrual age).  
Thickness and surface area maps were retrieved from automatically processed T2-weighted images \cite{makropoulos2018developing} and were parcellated into 34 bilateral cortical regions using the Desikan-Killiany-compatible M-CRIB-S(DK) atlas \cite{adamson2020parcellation}. 
For dHCP, we selected 25\% term-born infants for testing, yielding a balanced test set of 92 preterm and 93 term infants. 
Of the remaining 282 term-born infants, we selected 252 for training and 30 for validation.

\subsection{Stochastic cortical self-reconstruction \label{sec:scsrmethod}}
The key idea of the proposed stochastic cortical self-reconstruction (SCSR) is to use the cortex information in the MRI scan to estimate a subject-specific healthy reference distribution  (see Fig. \ref{fig:overview}). 
Based on a cortical thickness map extracted with FreeSurfer, SCSR randomly samples a fraction $s$ of vertex locations as predictors. A model is trained on healthy subjects from the UKB dataset to reconstruct the thickness values of the vertices that have not been sampled. As the model is only trained on healthy subjects, this can be thought of as a completion of the thickness map as if the subject were healthy. During the inference phase, we repeat the stochastic selection of vertices as input and healthy reconstruction $m$ times, yielding multiple partial reconstructions, see Fig. \ref{fig:overview}(b). From the resulting distribution of predicted thickness values at each vertex, an explicit healthy reference is obtained by selecting the values at a given centile $q$. Similar to established normative models, a Z-score is then computed by subtracting the actual thicknesses and dividing by the standard deviation estimated on the residuals of the UKB validation set. The resulting deviation maps highlight atypical thickness.

\begin{algorithm}[ht!]
\caption{Stochastic Cortical Self-Reconstruction \label{alg:scsr}}
\label{alg:dpp}

\SetKwBlock{TrainingPhase}{Training Phase}{}
\TrainingPhase{
    \KwIn{Training data $\mathcal{X} \in \mathbb{R}^{n \times p}$, sampling rate $s$}
    Initialize model $\mathcal{M}$\;
    \For{$i = 1$ to $n$}{
        Select subject $X \leftarrow \mathcal{X}_i$\;
        Stochastic sampling of $s$ features from $X$ as predictors $X_{\text{pred}}$\;
        Use the remaining features as responses $X_{\text{res}}$, s.t., $X = X_{\text{pred}} \cup X_{\text{res}}$\;
        Update $\mathcal{M}$ by minimizing the prediction error $\| X_{\text{res}} - \mathcal{M}(X_{\text{pred}}) \|^2$\;
    }    
    \Return $\mathcal{M}$
}

\vspace{1em}

\SetKwBlock{TestingPhase}{Testing Phase}{}
\TestingPhase{
    \KwIn{Test data $Y \in \mathbb{R}^{p}$, number of iterations $m$, reconstruction centile $q$, sampling rate $s$, model $\mathcal{M}$}
    Initialize tensor $T$ of size $p \times m$ with NaN values\;
    \For{$i \leftarrow 1$ \KwTo $m$}{
        Stochastic sampling of $s$ features from $Y$ as predictors $Y_{\text{pred}}$ \;
        Apply the trained model and store predictions: $T[:,i] = \mathcal{M}(Y_{\text{pred}})$\;
    }
    Reconstruct reference $R = \text{centile}(T, q)$ ignoring NaN values\;
    \Return $R$%
}

\end{algorithm}

Algorithm \ref{alg:scsr} details the training and testing procedure of SCSR. 
We consider $p$ cortical thickness values for $n$ healthy subjects in the training set $\mathcal{X} \in \mathbb{R}^{n,p}$ and a test subject $Y \in \mathbb{R}^{p}$. 
For the stochastic cortical self-reconstruction, we randomly sample $s$ thicknesses as predictors $X_{\text{pred}}$ and the remaining ones as responses $X_{\text{res}}$. 
The model $\mathcal{M}$ is optimized to minimize the prediction error. 
The model trained on healthy subjects is then applied to healthy or diseased subjects in the testing phase to predict the missing thicknesses $\mathcal{M}(Y_{\text{pred}})$. 
Repeating this process $m$ times yields $m$ partial reconstructions, where we reconstruct the healthy reference at a given centile~$q$. %

\begin{figure}[t]
    \centering
    \includegraphics[width=0.8\textwidth]{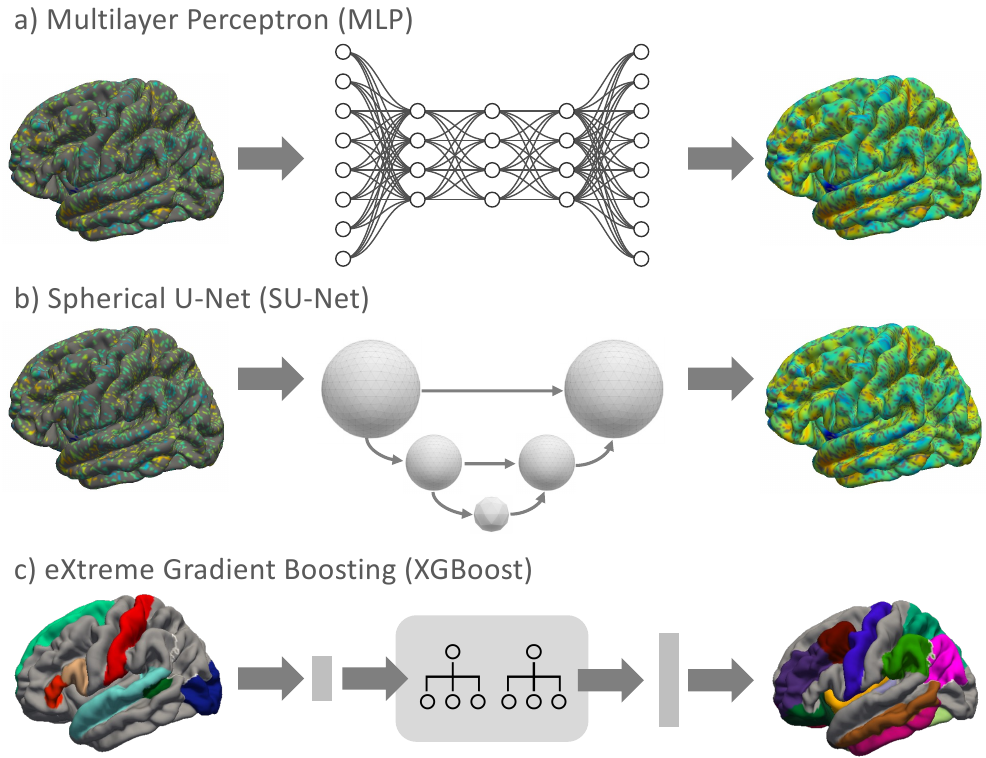}
    \caption{SCSR variants. MLP and SU-Net operate on vertex level, XGBoost on parcels. 
    (a) Thicknesses are vectorized, input to an MLP with three hidden layers, and the output is mapped back to the surface. 
    (b) Spherical U-Net uses spheres with cortical thickness values. 
    (c) The thicknesses of selected parcels serve as input to XGBoost, and the thickness values of the remaining parcels are predicted. }
    \label{fig:SUnet}
\end{figure}

\subsubsection{Implementations of SCSR \label{sec:implementations}}
Algorithm 1 describes the general concept of cortical self-reconstruction, where we implemented three versions (Fig. \ref{fig:SUnet}): two with deep neural networks and one with decision trees. 

\paragraph{Multilayer perceptron} 
The standard SCSR implementation used in our experiments is based on a multilayer perceptron (MLP). 
We used an autoencoder architecture with input and output dimensions corresponding to the resolution of the icosahedron, $p=10,242$. 
We chose a flat autoencoder design with three latent layers with 1,024 dimensions since we were not interested in the bottleneck latent space. 
Batch norm layers with Swift activation were placed between layers and a dropout layer with a rate of 0.5 at the second layer.
AdamW is used for optimization with a learning rate of 0.001, weight decay of 0.005, and 200 epochs. 
During training, a vertex sampling rate of 20\% was selected.
The value of non-sampled vertices was set to zero, corresponding to the mean, as we applied the standard scaler to the input data, subtracting the mean per feature. 
The loss function computes the mean squared error (MSE) between the predicted thickness and the actual thickness, where the loss is only computed on the non-sampled region. 
The model is trained for 200 epochs, and the checkpoint with the lowest reconstruction error on the UKB validation set is selected. 
\revision{Fig.\ref{fig:ukb_train_error} illustrates the mean absolute error on the training set of the selected model.}

\paragraph{Spherical U-Net}
Instead of flattening all vertices as for the MLP, we consider the spherical geometry of the cortex for the reconstruction by designing an autoencoder based on the spherical U-Net \cite{zhao2019spherical}. 
Fig. \ref{fig:SUnet} illustrates the spherical U-Net (SU-Net), which adapts the traditional U-Net architecture to handle spherical data. 
It is designed for the icosahedron used in FreeSurfer and uses the hierarchical expansion procedure of the icosahedron for up- and downsampling. 
The convolution layers of the spherical U-Net perform spherical convolutions and, therefore, learn spatial neighborhood relationships on the cortex. 
We use a spherical U-Net with 5 levels and a channel dimension of 16 at the highest level (doubling at each lower level). 
The training setting is identical to MLP. 

\paragraph{Parcellation-based SCSR with XGBoost \label{sec:parcelXGB}}
We also considered a parcel-based implementation of the SCSR concept, see Fig. \ref{fig:SUnet}, as prior normative models focused on parcels \cite{rutherford2022charting}. 
We used the Desikan-Killiany atlas \cite{desikan2006} in FreeSurfer to compute regional cortical thicknesses. 
We used eXtreme gradient boosting (XGBoost) \cite{chen2016xgboost} as model $\mathcal{M}$ for the prediction, an efficient implementation of gradient-boosted decision trees, and among the best-performing methods for tabular data. 
As XGBoost does not perform multivariate prediction, a separate XGBoost model is trained for each response variable in $X_{\text{rep}}$. 
This procedure is repeated $m$ times to obtain the stochastic self-reconstruction. 
In our experiments, we used a squared error regression objective with a learning rate of 0.15 and a maximum tree depth of 4.

\subsubsection{Atypical thickness deviation maps}
After training the model on the self-reconstruction task, we applied the model to the test data. 
As for the training, we sampled parts of the cortex as input, where we evaluated sampling rates from 1\% to 30\%. %
We repeated the reconstruction $m = 500$ times for each input. 
As the model is only trained on healthy subjects, this creates a distribution over healthy cortex reconstructions for that scan. 
We obtained an explicit healthy reference by selecting thickness values at centile $q$ at each vertex. 
The median, $q=0.5$, is a robust choice. 
However, pathologic brain changes are often related to neuron loss and decreased cortical thickness. 
If cortical regions affected by atrophy are sampled as predictors, this may lead to responses with lower thickness. 
To compensate for this in the reference creation, we also considered higher reconstruction centiles $q$ in \revision{our adult experiments with the standard setting of $q=0.95$}. 
Fig. \ref{fig:recons} illustrates reconstructions of healthy references for various centiles in the presence of dementia. 
Finally, we obtained cortical deviation maps by computing the vertex-wise Z-score, $Z = (Y - R) / \sigma $, with the actual thickness of test subject $Y$, SCSR reference $R$, and standard deviation $\sigma$, estimated from residuals of the UKB validation set. 

\subsubsection{SCSR modifications \label{sec:variations}}
SCSR exclusively uses cortical thickness values as input, unlike common normative models that use age and sex. 
We also considered adding age and sex as additional input to SCSR. 
Given our use of an MLP, these variables could be easily concatenated with the existing input.
Furthermore, we evaluated two variations of the stochastic sampling of vertex values. %
The first variation leveraged the Desikan-Killiany parcellation by sampling all vertices in a parcel instead of individual vertices. 
This created a more challenging learning task, as the model must operate with less granular information. 
\revision{The second modification involved excluding predictors from cortical regions associated with AD from training and inference.} The rationale is that the selected predictors are used to create a healthy reconstruction; including atrophied vertices might bias the model toward a diseased representation. Since AD-affected regions have been previously identified in the AD-ROI \cite{schwarz2016large} (Fig. \ref{fig:AD-ROI}), we used this information to avoid sampling from those areas.
\revision{Finally, we also consider the harmonization of cortical thickness values before inputting them into SCSR via ComBat, which uses an empirical Bayes framework to remove site-related batch effects \cite{fortin2018harmonization,wachinger2021detect}.}

\subsection{Alternative reference models \label{sec:baselines}}
\textbf{Adults:} Normative models are the common choice for creating healthy references, using age and sex as input \cite{rutherford2022charting,rutherford2022normative}. 
We compare to a generalized additive model (GAM), specified as  
$Y \sim \alpha_0 + f(\text{age}) + \alpha_1 \cdot (\text{sex})$, where $f$ is a smooth spline function estimated based on the data to model the non-linear relationship between thickness and age. 
Further, we compare to GAMLSS that considers next to the mean (location) also the variance (scale), skewness, and kurtosis (shape) of the distribution of the response variable.
GAMLSS has previously been used for estimating normative growth curves of cortical thickness, where we specify the first, second, and third order moments similar to (\cite{bethlehem2022brain}, Eq.8) considering coefficients $\alpha, \beta, \gamma$:
\begin{align}
\log(\mu) &\sim \alpha_0 + \alpha_1 \cdot (\text{age})^{-2} + \alpha_2 \cdot (\text{age})^{-2} \log(\text{age}) + \alpha_3 \cdot (\text{sex}), \\
\log(\sigma) &\sim \beta_0 + \beta_1 \cdot (\text{age})^{-1} + \beta_2 \cdot (\text{age})^{0.5} + \beta_3 \cdot (\text{sex}),
\end{align}
and $\nu \sim \gamma_0$. 
For GAMLSS, we directly predicted the standard deviation for the Z-score computation from the model. 
GAM and GAMLSS models were estimated on the UKB training data. 
Additionally, we included a Bayesian linear regression (BLR) model implemented in the PCNtoolkit \cite{rutherford2023evidence} to  predict regional thickness from the covariates age and sex
Briefly, for each brain region, thickness is modeled as an estimated weight vector multiplied by a basis expansion of the covariate vector, consisting of a B-spline basis (cubic spline with five evenly spaced knots) to model non-linear effects of age. Non-Gaussian effects were modeled with likelihood warping \cite{fraza2021warped}.
We considered the pre-trained model \texttt{lifespan\_29K\_82sites} and refitted it to the UKB data, as well as, training a BLR model directly on the UKB data. 

These prior reference methods operate at the parcel level. 
To also have a baseline method that operates on the vertex level,  we implemented a reference population model (Pop-Ref) based on age brackets. 
To this end, we defined age brackets of 5 years on the UKB training data. 
For each bracket, we estimate the vertex-wise mean and standard deviation. 
For a test subject, we select the age bracket based on the actual age and then compute the Z-score. %

\textbf{Infants:} For infants of the dHCP cohort, normative reference charts for bilateral regional cortical thickness (CT) were obtained from the BrainChart project \cite{bethlehem2022brain}, in which GAMLSS-based normative models were computed over the human lifespan as a function of age and sex from about 100,000 MRI scans. To adapt the reference charts to dHCP, random effects of the study site were calculated based on term-born individuals. 
A deviation score was then calculated for each individual and each cortical region, which represents the percentile of this person's CT value for the respective cortical region in relation to the normative range. 
As a control analysis, two different approaches for obtaining normative ranges of regional CT were investigated, i.e., quantile regression and Bayesian linear regression. For quantile regression, regional CT quantiles were modeled as a function of age and sex using the \texttt{quantreg} function in Matlab 2023b. 
Following a previous study \cite{di2022cell}, term-born individuals from the dHCP training set were used for training, and then deviations from the normative range were calculated for each cortical region as a Z-score. 
For BLR, we could not use pretrained models released previously with the PCNtoolkit \cite{rutherford2022charting} because they do not include infants before the age of 2 years. 
Instead, we estimated the BLR model on the dHCP training set. 
Next, thickness deviations were estimated as Z-scores for each brain region. 
For both control approaches as well as for SCSR, Z-scores were converted into centile scores using the cumulative distribution function of the standard normal distribution to facilitate a comparison with the centile scores computed with the GAMLSS-based normative model. The median difference between centiles across groups was used to compare the estimations across approaches. Significance was assessed with a permutation test (n=10,000; percentileTest of rcompanion package in R).

\subsection{Evaluation metrics \label{sec:metrics}}
The Receiver Operating Characteristic Area Under the Curve (ROC AUC) was used to assess the model's ability to distinguish between classes. The ROC curve plots sensitivity (true positive rate) against 1-specificity (false positive rate) at various threshold settings, providing a comprehensive measure of classification performance. Instead of applying it directly to model outputs, we applied it to Z-scores to evaluate their discriminative power. Given that each vertex has a Z-score, we computed the mean Z-score in the AD ROI for AD experiments, and across the entire cortex for differential diagnosis experiments. Since ROC AUC is designed for binary classification, we handled the three-class (AD, MCI, CN) problem by calculating the ROC AUC for each binary task and averaging the results.
For the Spearman correlation, we encoded diagnoses as ordinal variables (0: CN; 1: MCI; 2: AD) and computed the correlation between these values and the Z-scores in the AD ROI to measure the association between disease progression and cortical thickness deviation.

Regarding \emph{reconstruction error}, we calculated the mean squared error (MSE) between actual and reconstructed cortical thicknesses. Errors were computed at both the parcel and vertex levels. For parcel-based models, we first averaged the vertex thicknesses within each parcel before calculating the MSE to ensure consistency with parcel-based reference models.

\section{Results}

\subsection{Identifying atrophy patterns across AD datasets} %

\begin{figure}
    \centering
    \includegraphics[width=0.8\linewidth]{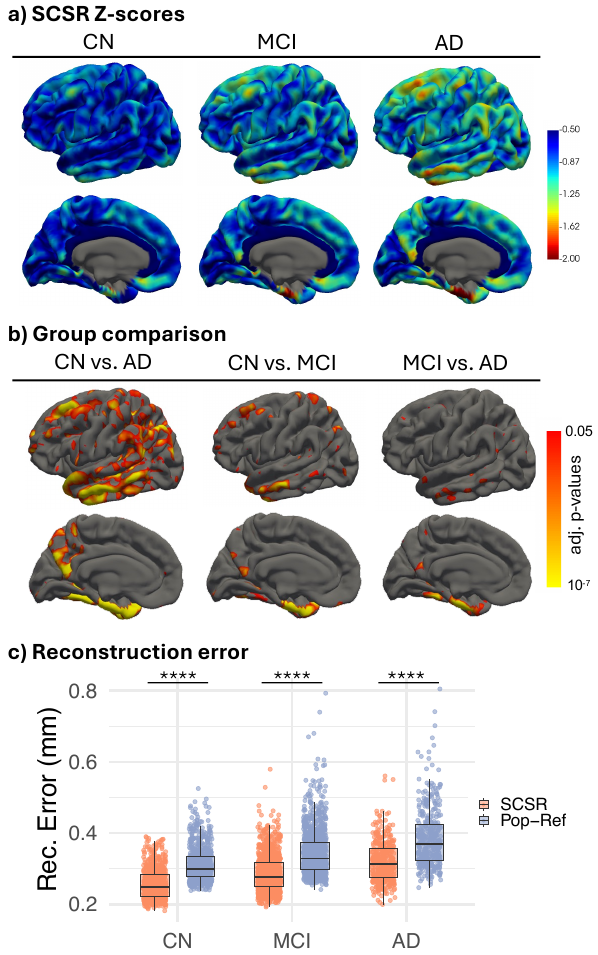}
    \caption{%
    (a) Vertex-wise Z-scores from SCSR for CN, MCI, and AD groups. (b) Adjusted $p$-values from group comparisons after Bonferroni correction (Wilcoxon rank-sum test) on log scale. 
    (c) Boxplot and jitter plot displaying the \revision{mean absolute reconstruction error (in mm) by subject across diagnostic groups}. SCSR is compared to Population Reference (Pop-Ref), and the statistical significance was evaluated with the Wilcoxon signed-rank test (****: $p < 0.0001$).
    Boxplots indicate the median and interquartile range, while whiskers extend to 1.5 times the interquartile range.
    }
    \label{fig:AD-summary}
\end{figure}

Detecting atrophy is crucial for diagnosing Alzheimer's disease and tracking the progression of cognitive decline \cite{frisoni2010clinical}. 
We estimated healthy references with SCSR on four datasets containing subjects with mild cognitive impairment (MCI) and AD: the Alzheimer’s Disease Neuroimaging Initiative (ADNI) \cite{jack2008alzheimer}, the Japanese ADNI (J-ADNI) \cite{iwatsubo2010japanese},  the Australian Imaging, Biomarkers and Lifestyle (AIBL) study \cite{ellis2009australian}, and the 
DZNE-Longitudinal Cognitive Impairment and Dementia Study (DELCODE) \cite{jessen2018design} (see Sec. \ref{sec:data} for data details).
These datasets come from four different continents and have not been used during SCSR training. 
We split the ADNI dataset into a 20\% validation set and an 80\% test set.
On the ADNI validation set, we compared different SCSR configurations and evaluated the impact of sampling rate and centile, see Sec. \ref{sec:ablation}. 
The ADNI results reported in this section are on the test set.

Fig. \ref{fig:AD-summary}(a) illustrates mean Z-scores from SCSR  for three groups: cognitively normal (CN), MCI, and AD. 
The lowest Z-scores, indicating the most significant atrophy, are observed in AD patients, particularly in the medial temporal lobe, temporoparietal junction, and posterior cingulate cortex. 
The detected atrophy pattern for MCI subjects is similar, though less pronounced, with the lowest Z-scores in the medial temporal lobe. 
CN subjects show only small, isolated areas of low Z-scores.
Fig. \ref{fig:AD-summary}(b) illustrates significant vertex-wise p-values after Bonferroni correction from pairwise group comparisons using the Wilcoxon rank-sum test on Z-scores. 
Highly significant differences for MCI subjects compared to CN and AD in the temporal region, which can support the early diagnosis. 
The comparison of CN and AD shows widespread atrophy, extending from the temporal to the parietal lobe and also in the frontal parts of the brain. 

Fig.\ref{fig:AD-summary}(c) plots \revision{reconstruction errors across the cortex between the original thickness and reconstructed healthy reference} for three diagnostic groups. 
We compare the reconstruction errors of SCSR with population references (Pop-Ref; Sec. \ref{sec:baselines}) as a vertex-wise baseline approach. 
SCSR yields significantly lower reconstruction errors than Pop-Ref (paired Wilcoxon signed-rank test). 
Moreover, the error increases from CN to MCI and further to AD, as expected, since SCSR constructs healthy references.

\begin{figure}[th!]
    \centering
    \includegraphics[width=0.99\linewidth]{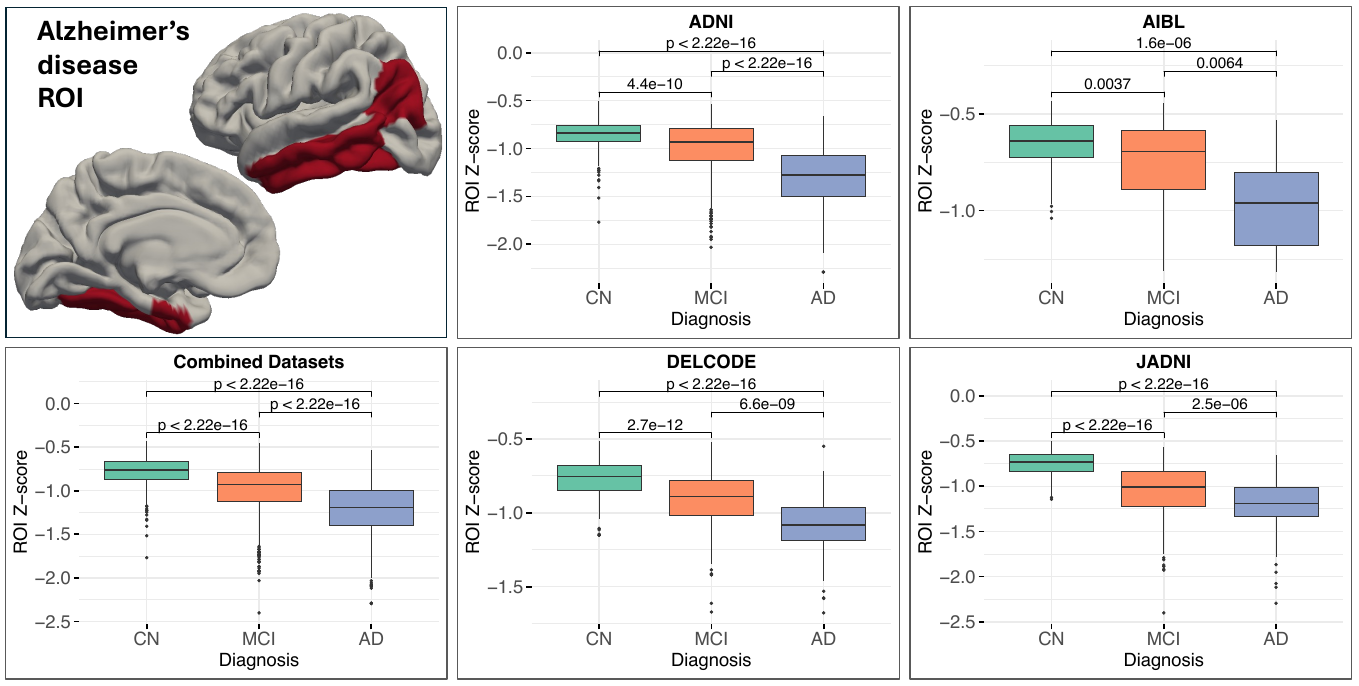}
    \leavevmode\vspace{0.2cm}
    \setlength{\tabcolsep}{2pt}
    \centering
    \resizebox{\linewidth}{!}{%
    \rowcolors{2}{gray!15}{white}
    \begin{tabular}{l *{10}{S}}%
    \toprule
    & \multicolumn{5}{c}{ROC AUC} & \multicolumn{5}{c}{Spearman correlation} \\ 
    \cmidrule(lr){2-6}
    \cmidrule(lr){7-11}
~ & \emph{All} & {ADNI} & {AIBL} & {DELCODE} & {JADNI} & \emph{All} & {ADNI} & {AIBL} & {DELCODE} & {JADNI}  \\       
        \midrule
        SCSR  & 0.80 & 0.77 & 0.74 & 0.80 & 0.81 & -0.54 & -0.42 & -0.34 & -0.57 & -0.59  \\ 
        GAM & 0.74 & 0.75 & 0.69 & 0.71 & 0.74 & -0.41 & -0.40 & -0.26 & -0.40 & -0.44  \\ 
        GAMLSS & 0.76 & 0.77 & 0.75 & 0.74 & 0.79 & -0.44 & -0.43 & -0.27 & -0.46 & -0.55  \\ 
        BLR  & 0.72 & 0.76 & 0.73 & 0.71 & 0.73 & -0.37 & -0.40 & -0.27 & -0.39 & -0.43  \\ 
        BLR pretr. & 0.75 & 0.76 & 0.74 & 0.72 & 0.75 & -0.43 & -0.40 & -0.28 & -0.43 & -0.47 \\ 
        Pop-Ref & 0.76 & 0.75 & 0.76 & 0.73 & 0.77 & -0.48 & -0.40 & -0.33 & -0.44 & -0.50  \\

    \bottomrule
    
    \end{tabular}
    }
    \vspace{0.3cm}
    \caption{Top: Illustration of AD ROI based on \cite{schwarz2016large} (left). Boxplots illustrate the distribution of ROI Z-scores across diagnostic groups and AD datasets. $p$-values are computed with the Wilcoxon rank-sum test. 
    Bottom: Comparison of SCSR with normative models for group discrimination quantified with Area Under the Receiver Operating Characteristic Curve (ROC AUC) and Spearman correlation. Results are reported for the combination of all AD datasets and each one separately. GAM: generative additive model; GAMLSS: GAM of location, scale, and shape; BLR: Bayesian linear regression; BLR petr.: BLR pretrained.}
    \label{fig:AD-ROI}    
\end{figure}

\revision{For quantifying the group differentiation of SCSR, we compute the average Z-score within an AD region of interest (ROI) that encompasses the previously reported \cite{schwarz2016large} five parcels: entorhinal, inferior temporal, middle temporal, inferior parietal, and fusiform cortices.}  
Fig. \ref{fig:AD-ROI} illustrates the ROI and boxplots of ROI Z-scores for all four datasets and the combined dataset.  
Highly significant group differences between diagnostic categories are observed across all datasets, as evidenced by the p-values from the Wilcoxon rank-sum test.
A consistent pattern emerges across all datasets, with the AD group exhibiting the largest deviation, followed by MCI and CN.
\revision{Note that the Z-scores of the CN group are not distributed around zero because of $q=0.95$, resulting in thicker healthy reconstructions.}
We further evaluated the group discrimination performance of SCSR using the Area Under the Receiver Operating Characteristic Curve (ROC AUC; Sec. \ref{sec:metrics}) based on ROI Z-scores. 
ROC AUC was computed for all pairwise combinations of diagnoses, and the mean values are presented in Fig. \ref{fig:AD-ROI} for each dataset and the combined dataset. 
Additionally, Spearman correlations between ROI Z-scores and the three diagnoses, encoded as an ordinal variable (details in Sec. \ref{sec:metrics}), are reported. 
The best discrimination based on ROI Z-scores is seen in DELCODE and J-ADNI, followed by ADNI and AIBL.

\begin{figure}[t]
    \centering
    \includegraphics[width=0.99\linewidth]{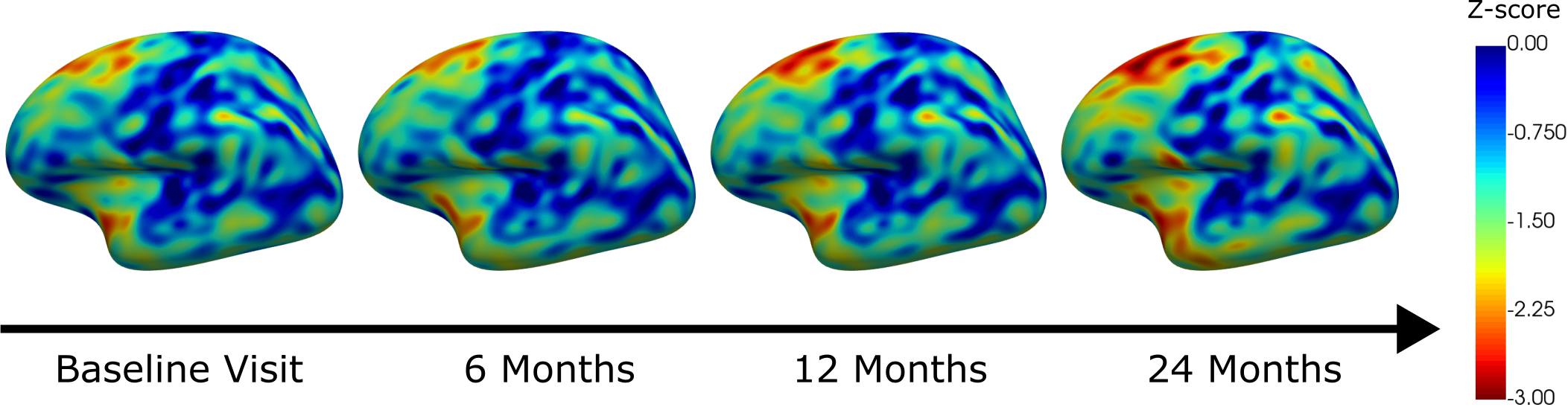}
    \caption{SCSR Z-score maps for a single subject (male, 76 years, AD dementia\revision{, ID 153\_S\_4172}) for baseline and follow-up visits. The maps were computed independently for each visit and reveal a consistent deviation pattern that becomes more abnormal in AD regions as dementia progresses. }
    \label{fig:ADlongitudional}
\end{figure}

For comparison, we evaluated the performance of normative modeling approaches operating on the parcel level, including the generative additive model (GAM), generalized additive models of location, scale, and shape (GAMLSS) \cite{rigby2005generalized,bethlehem2022brain,dinga2021normative}, and Bayesian linear regression (BLR) \cite{fraza2021warped}, as implemented in the PCNtoolkit \cite{rutherford2022normative}. 
All models use the same UKB training set for estimating coefficients; for BLR, we also adopted a pretrained model (BLR pretr.). 
As an alternative vertex-wise model, we included age-stratified population references (Pop-Ref), with details of these approaches reported in Sec. \ref{sec:baselines}.  
The results in Fig. \ref{fig:AD-ROI} demonstrate SCSR offers the highest AUC values and strongest correlations. 

To assess the consistency of individual Z-score maps, Fig. \ref{fig:ADlongitudional} illustrates SCSR results for an AD patient with multiple follow-up visits.
For each visit, the Z-score map is computed independently from the SCSR reconstruction. 
The maps exhibit a highly consistent pattern across visits. 
Together with previous results on low reconstruction errors, this further validates the estimation of subject-specific healthy references using SCSR. 
Additionally, a decrease in Z-scores over successive visits suggests progressive atrophy in characteristic AD regions, highlighting the potential of SCSR for tracking disease progression.

\subsection{Differential diagnosis of dementia using routine clinical data}
Accurate differential diagnosis of dementia is critical for guiding treatment and patient management. Previous studies have identified characteristic cortical atrophy patterns associated with different types of dementia \cite{du2007different,chouliaras2023use}. 
We assessed the ability of SCSR to detect distinct cortical atrophy patterns by analyzing routine clinical data from 50 subjects in our hospital database. 
SCSR was applied to each individual scan without any dataset-specific adaptation, presenting a realistic scenario of clinical translation. 
The dataset included 10 patients per group, covering AD, posterior cortical atrophy (PCA), behavioral variant frontotemporal dementia (bvFTD), semantic dementia (SD), and cognitively normal (CN) controls.

\begin{figure}
    \centering
    \includegraphics[width=\textwidth]{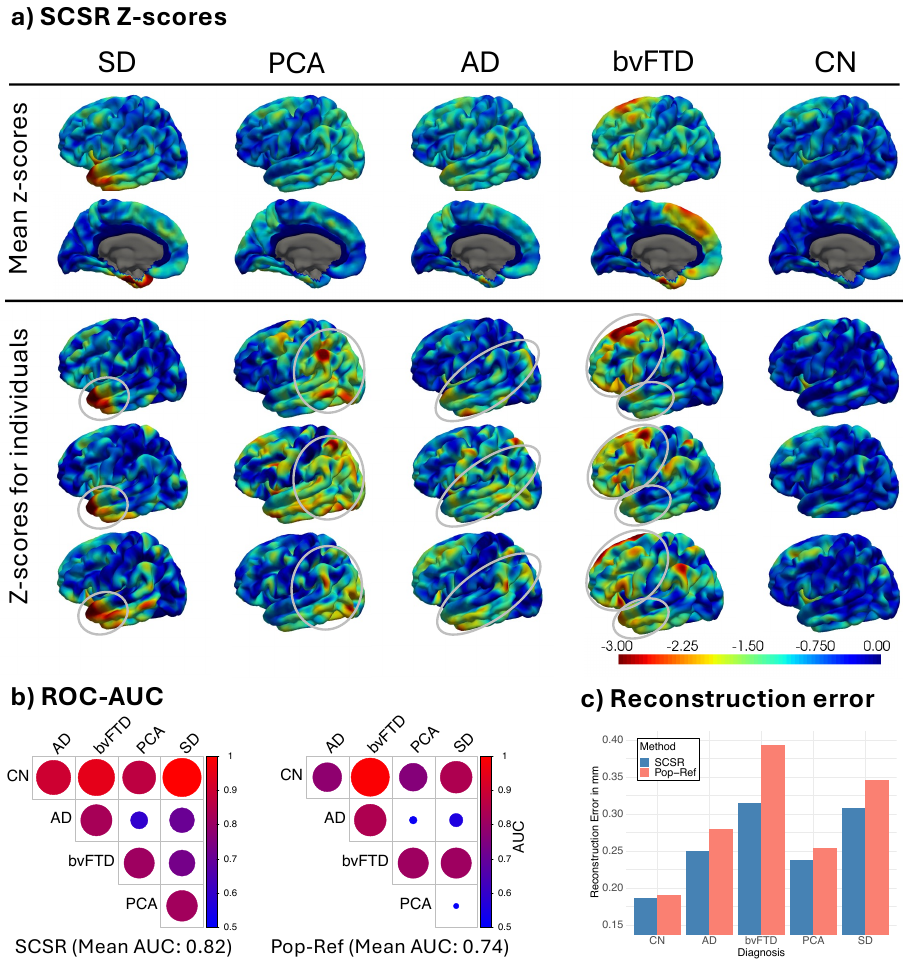} 
    \caption{Differential diagnosis of dementia on routine clinical data, covering Alzheimer's disease (AD), posterior cortical atrophy (PCA), behavioral variant frontotemporal dementia (bvFTD), semantic dementia (SD), and cognitively normal (CN). (a) Mean and individual Z-scores across diagnostic groups. Ellipses highlight regions with the strongest atrophy. Medial views are illustrated in Fig. \ref{fig:diffDiag_supp}. (b) Pairwise discrimination of groups by computing the ROC-AUC on Z-scores averaged across the cortex. \revision{Circle size and color are identical for both plots.} SCSR demonstrated better discrimination than Pop-Ref. (c) Reconstruction error across all diagnostic groups for both approaches.}
    \label{fig:diff-diag}
\end{figure}

Fig. \ref{fig:diff-diag}(a) visualizes vertex-wise average Z-scores from SCSR for the four dementia groups and CN in the clinical dataset.  
In all four diagnostic patient groups, we see localized areas with highly negative Z-scores, indicating cortical atrophy in these areas.
For bvFTD patients, SCSR indicates lower thickness in prefrontal regions and lateral temporal cortical areas, which is considered a hallmark finding on MRI in these patients. 
Fig. \ref{fig:diff-diag}(a) illustrates Z-score maps of individual subjects, demonstrating the ability to highlight individual deviations. 
As indicated by its name, posterior cortical atrophy is characterized by atrophy in parietal and occipital cortical areas, highlighted with the strongest deviations.
We appreciate abnormal thickness in the temporoparietal junction in AD, as well as in the posterior cingulate and mesial temporal lobe. 
SD leads to a loss of semantic memory and language comprehension, with an expected atrophy in the anterior temporal lobes. 
Overall, SCSR indicates dementia-specific areas of regional atrophy in a clinical patient cohort aligning with established neuroimaging findings~\cite{du2007different,risacher2013neuroimaging}. 

We use the AUC ROC to quantify the ability of Z-scores to discriminate between all diagnosis pairs (Fig. \ref{fig:diff-diag}(b)). Unlike our previous AD experiments, we compute the Z-scores across the entire cortex rather than focusing on the AD ROI. Compared with Pop-Ref, SCSR demonstrates an improvement in AUC from 0.74 to 0.82, with the most notable gains observed in distinguishing between AD-PCA, AD-SD, and PCA-SD pairs.
Additionally, Fig. \ref{fig:diff-diag}(c) presents the reconstruction error for each diagnostic group, with the lowest error seen in CN and the highest in FTD. Across all groups, SCSR outperforms Pop-Ref by yielding lower reconstruction errors. This indicates that SCSR enables better discrimination among dementia types and provides more accurate estimations of cortical thickness references.
When comparing the reconstruction errors among the four types of dementia, we can appreciate their different magnitudes of cortical pathology.  
In our sample, the magnitude is highest for bvFTD, followed by SD, AD, and PCA. 
As the reconstruction error summarizes the cortical thickness deviation from the healthy reference, it represents an aggregate atypicality of an individual scan. 

To further illustrate how SCSR constructs healthy reference maps, Fig. \ref{fig:recons} visualizes reference thickness maps for different centiles $q$. These references not only resemble the actual cortical thickness but also show subject-specific variations, highlighting the individualized nature of the references. %

\subsection{Atypical cortical thickness in preterm infants}

Preterm infants, born before 37 weeks of gestation, are at higher risk for developmental delays and health complications due to various dysmaturational disturbances in the brain \cite{volpe2019dysmaturation}. 
Early interventions and monitoring are critical for improving long-term outcomes in this population. 
The developing Human Connectome Project (dHCP) collected MRI data of term and preterm-born infants at term-equivalent age (see Sec. \ref{sec:data} for details on data processing and splits). 
Normative modeling is challenging in this scenario because of limited brain imaging data of infants \cite{bethlehem2022brain}. 
Additionally, the mapping of age to cortical thickness is confounded by early birth, occurring during a period of rapid brain development around the time of birth.

\begin{figure}[ht!]
    \centering
    \includegraphics[width=0.99\linewidth]{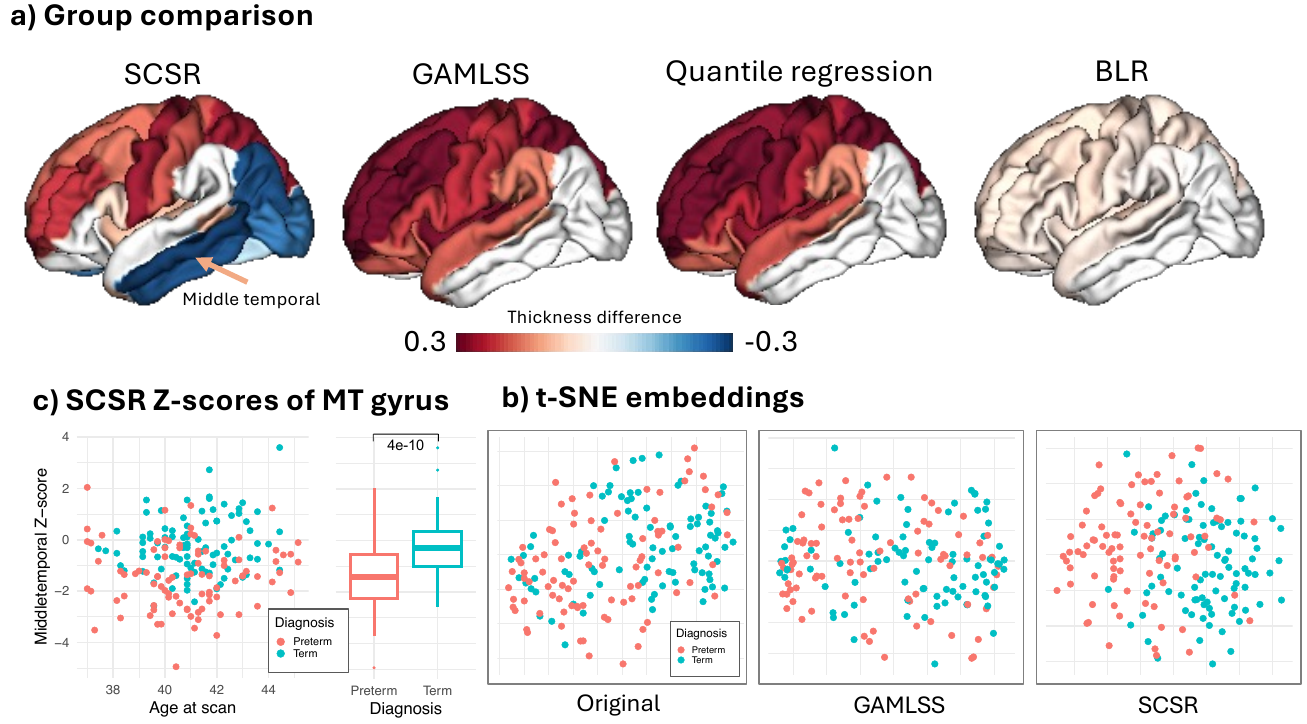}
    \caption{Atypical cortical thickness patterns in preterm infants at term-equivalent age. (a) Only significant group differences between term and preterm infants are shown in color, with the color corresponding to the difference in median thickness; blue indicates lower cortical thickness in the preterm group. Comparison of SCSR with normative models (GAMLSS, quantile regression, BLR) shows atypical thinning detected by SCSR. Medial views in Fig. \ref{fig:preterm_full}. (b) t-SNE embeddings into 2D of all dHCP test subjects based on the original thickness, GAMLSS, and SCSR, with the SCSR embedding showing \revision{a smaller group overlap}. (c) SCSR Z-scores of the middle temporal (MT) gyrus across age at scan and groups.  }
    \label{fig:preterm}
\end{figure}

Fig. \ref{fig:preterm}(a) highlights cortical regions with significant group differences in cortical thickness centile estimates (permutation test with 10,000 permutations; percentileTest of the rcompanion package; p\textsubscript{FDR}<0.05). 
For comparison, we used GAMLSS-based reference charts from the BrainChart project \cite{bethlehem2022brain}, quantile regression \cite{di2022cell}, and Bayesian linear regression (BLR) from PCNtoolkit \cite{rutherford2022normative}, detailed in Sec. \ref{sec:baselines}. 
For consistency with these parcel-based approaches and the limited sample size of dHCP, we used the parcel-based variant of SCSR based on XGBoost in this experiment with centile $q=0.5$ (see Method \ref{sec:implementations}). 
Unlike our other experiments, where the UKB dataset was used for training, this variant was trained on dHCP data.
For comparability, the same dHCP training set was also used for quantile regression and BLR. In the case of GAMLSS, the training set was used to estimate site-specific offsets.

SCSR identified significantly lower cortical thickness for preterms in the middle temporal, inferior temporal, inferior parietal, and lateral occipital regions. 
Contrastively, the three normative models do not show significant group differences for these regions, which might be a relevant finding since lower thickness in these regions is evident in later developmental stages \cite{kelly2024cortical,schmitz2020decreased}. 
The middle temporal (MT) gyrus showed the largest difference between groups in cortical thickness. 
Fig. \ref{fig:preterm}(c) plots MT Z-scores across age at scan for both groups and summarizes their distribution with boxplots, highlighting the significantly lower individual thickness centile estimates for preterm infants. 
Finally, Fig. \ref{fig:preterm}(b) visualizes 2D embeddings from t-SNE by using data from all parcels as input. Specifically, we use the original cortical thickness values, as well as deviations from GAMLSS and SCSR. 
The embeddings on the original and GAMLSS-derived data do not indicate a group separation, while the SCSR embedding shows a separation of both groups. 
As t-SNE is an unsupervised approach, this indicates that SCSR effectively captures structural differences between the groups across the cortex.

\subsection{Insights into design choices of SCSR \label{sec:ablation}}

\begin{figure}
    \centering
    \begin{subfigure}{\textwidth}

    \resizebox{\textwidth}{!}{
    \rowcolors{4}{gray!15}{white}
    \begin{tabular}{ccccccc}
    \toprule
        ~ & ~ & ~ & ~ & ~ & \multicolumn{2}{c}{Rec Error}    \\ 
        \cmidrule(lr){6-7}
        Parcel sampling & Ignore AD regions  & Age \& Sex & Harmonization & AUC & CN & AD  \\ 
        \midrule
        
        \checkmark & \xmark & \xmark & \xmark & 0.898 & 0.178 & 0.263  \\ 
\xmark & \checkmark & \xmark & \xmark & 0.892 & 0.184 & 0.263  \\ 
\xmark & \xmark & \checkmark & \xmark & 0.884 & 0.155 & 0.220 \\         
\xmark & \xmark & \xmark & \checkmark & 0.895 & 0.181 & 0.253  \\ 
\xmark & \xmark & \xmark & \xmark & 0.901 & 0.185 & 0.254  \\ 
        \bottomrule
    \end{tabular}
    }
        \caption{AUC and reconstruction error (parcel level) for different SCSR configurations. %
        \label{tab:ablation}}
    \end{subfigure}
    
     \begin{subfigure}{\textwidth}
    \includegraphics[width=0.99\linewidth]{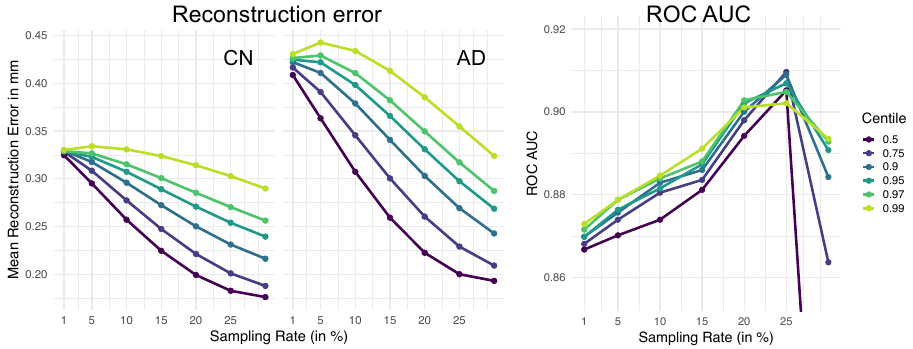}
    \caption{AUC and reconstruction error (vertex level) for varying sampling rates $s$ and reconstruction centiles $q$. %
    \label{fig:rec_roc_validation}}
    \end{subfigure}

    \caption{Evaluation of different SCSR configurations and parameters on the ADNI validation set. Reconstruction error (in mm) is reported for CN and AD; ROC AUC for the differentiation of both groups based on mean Z-scores in the AD ROI. }
\end{figure}

While the core idea of SCSR is to learn about cortical relationships and to use this for repeated, stochastic self-reconstruction, the concrete implementation requires several design choices (details in Sec. \ref{sec:variations}). 
Fig. \ref{tab:ablation} reports modifications of SCSR with results performed on the ADNI validation set, which consists of 20\% randomly sampled CN and AD subjects from ADNI. 
First, we used a parcel- instead of a vertex-wise sampling strategy, which presents a more difficult training task, as there are larger gaps with missing information that need to be reconstructed. 
Second, we avoided sampling from the AD ROI (illustrated in Fig. \ref{fig:AD-ROI}) as the sampling of areas with atrophy may prevent the reconstruction of a healthy reference. 
Third, we included age and sex as additional inputs to SCSR, conforming to existing normative modeling approaches. 
Fourth, we harmonized the thickness values across UKB and ADNI with ComBat \cite{fortin2018harmonization,wachinger2021detect}. %
\revision{None of these modifications improved the AUC, but the inclusion of age and sex lowered the reconstruction error. 
It did not translate to a higher AUC because effective discrimination requires low error in CN but relatively high error in AD.}

Furthermore, we investigated two of SCSR's parameters: sampling rate $s$ and reconstruction centile $q$. 
\revision{We varied the sampling rate during inference, while we kept it at 20\% during training.}
Fig. \ref{fig:rec_roc_validation} illustrates reconstruction errors for CN and AD subjects, together with the AUC on the ADNI validation set. 
The reconstruction error increases as sampling rates decrease and centiles rise.
Additionally, the AD group consistently exhibits higher reconstruction errors than the CN group across all configurations, reinforcing that SCSR effectively generates healthy references.
The AUC reaches its highest values at sampling rates between 20\% and 25\%. 
\revision{When the sampling rate exceeds 30\%, the reconstruction is too similar to the input, prohibiting the reconstruction of a healthy reference for AD, and thus yielding a reduction in AUC.} 
Furthermore, a general trend suggests that higher reconstruction centiles lead to slightly improved AUC.
As a trade-off, we selected a sampling rate of 20\% and the 95th centile as the standard configuration used throughout all adult experiments. 
Fig. \ref{fig:recons} illustrates SCSR reconstructions at different centiles and their corresponding Z-score map for individual patients.

Finally, we have explored different architectures for implementing SCSR (see Sec. \ref{sec:implementations} and Fig. \ref{fig:SUnet}). 
MLPs are flexible neural networks that do not pose spatial constraints on the input data, treating all thickness values as a high-dimensional vector. 
In contrast, the spherical U-NET (SU-Net) \cite{zhao2019spherical} was proposed for processing cortical surfaces, and it implements convolutions on the sphere. 
We used FreeSurfer for mapping cortical thicknesses to the sphere and then reconstructed surface data with SU-Net. 
Additionally, we implemented a version of SCSR that operates on the parcel level and predicts with extreme gradient boosting (XGBoost). 
This approach significantly reduces the input dimensionality compared to vertex-level methods, simplifying the training process. However, this reduction in complexity comes at the cost of capturing fine-grained cortical relationships. Table \ref{tab:architectures} reports the performance of the three architectures on the test dataset, with MLP achieving the best results, closely followed by SU-Net, and then XGBoost.

\begin{figure}
    \centering
    \includegraphics[width=\linewidth]{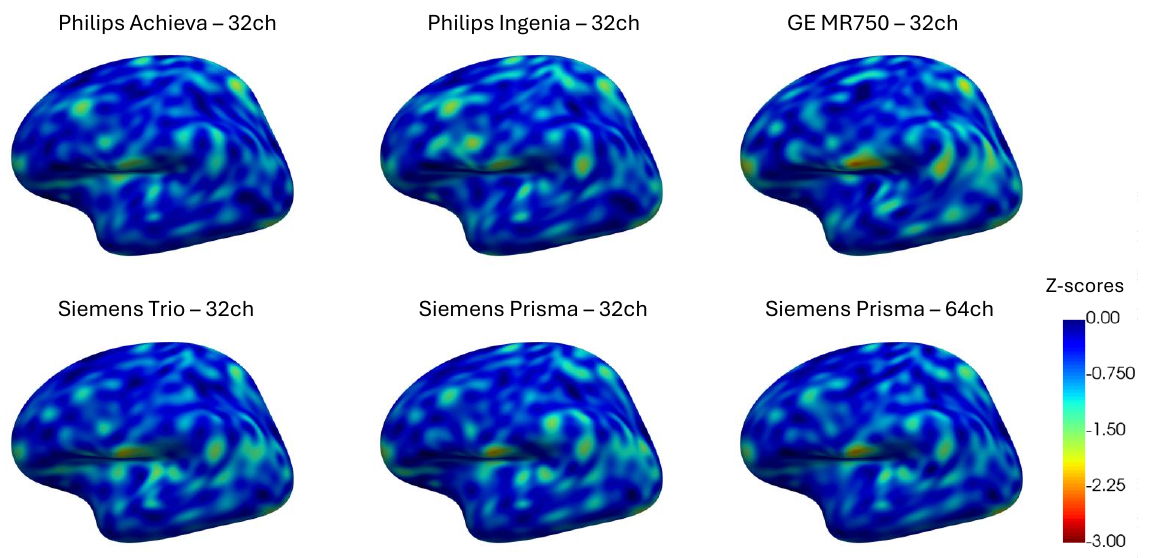}
    \caption{\revision{Robustness to scanner-related variability of SCSR generated Z-score maps. The same subject (male, 48 years, healthy, ID 03286)  was scanned with five different MRI scanners from three vendors (GE/Philips/Siemens) and two different coils (32 and 64 channels).}}
    \label{fig:scanner-robustness}
\end{figure}

\begin{figure}
    \centering
    \includegraphics[width=0.4\linewidth]{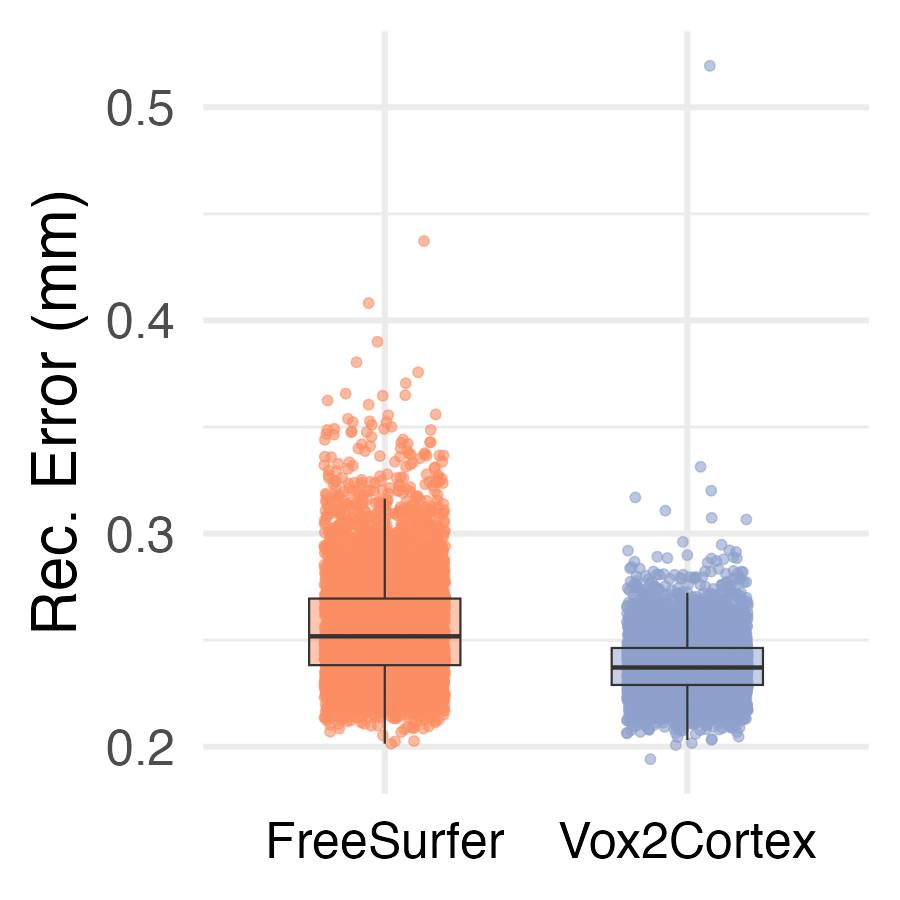}
    \caption{\revision{Boxplot and jitter plot displaying the mean absolute reconstruction error (in mm) of SCSR on the UKB validation set. Comparison of FreeSurfer and Vox2Cortex-Flow for the estimation of cortical thickness maps from the MRI.}}
    \label{fig:v2c}
\end{figure}

\subsection{\revision{SCSR robustness}}
\revision{
To assess the robustness of SCSR to scanner-related image variability, we used a recent traveling-heads dataset~\cite{Warrington2025onharmony}. We arbitrarily selected the first subject of the dataset and generated Z-score maps with SCSR for each available scan, as shown in Fig.~\ref{fig:scanner-robustness}. Across all scanner types, SCSR produced highly consistent outputs, preserving the subject-specific pattern of deviations. Minor variations in Z-scores were observed in the occipital lobe scanned with the GE MR750 scanner; however, these differences did not compromise the overall interpretation of the subject exhibiting a typical cortical thickness pattern. These results indicate that SCSR exhibits strong robustness to inter-scanner variability, supporting its applicability in multi-site neuroimaging studies.
}

\revision{
All previous results were based on cortical thickness extracted by FreeSurfer, spherically registered, and resampled to the FsAverage icosphere. %
However, recent advances in deep learning-based cortical surface reconstruction have introduced new methods that offer substantially reduced processing times while maintaining anatomical accuracy. To evaluate the impact of using a different reconstruction pipeline, we trained an SCSR model, identical in architecture and training configuration, on UKB data processed with Vox2Cortex-Flow~(V2C-Flow)~\cite{Bongratz2024v2cflow}.
On the UKB validation set, the SCSR model trained on V2C-Flow data achieved a lower reconstruction error (mean: 0.238 mm, SD: 0.014 mm) compared to the model trained on FreeSurfer data (mean: 0.256 mm, SD: 0.024 mm), see also Fig.~\ref{fig:v2c}. Importantly, we used the raw output of V2C-Flow, which directly provides correspondences without the need for explicit spherical registration. 
These results suggest that SCSR generalizes well across different cortical reconstruction methods, %
and may even benefit from the surface data offered by modern deep learning-based pipelines, whose higher efficiency strengthens clinical applicability. %
}

\section{Discussion}

SCSR introduces a novel approach to generating healthy cortical references by relying solely on the individual's own cortex as input. 
Across multiple experiments, we have demonstrated that SCSR provides accurate identification of brain pathology and lower reconstruction errors compared to existing reference models. %
Current normative models explicitly map biological variables such as age and sex to generate a healthy cortical reference. However, these variables provide only a rough approximation of an individual’s characteristics, and adding factors like education, ethnicity, and socioeconomic status complicates their application, as these variables must be consistently defined and available for all subjects. 
Additionally, our experiments on preterm infants indicated the limitations of age as a predictor, where it can become confounded. In contrast, SCSR constructs a healthy reference by learning implicit cortical relationships, avoiding the need for such external variables and their associated complexities.
In fact, adding these variables as additional input to SCSR did not improve AUC, suggesting that information from the cortex itself provides the most accurate information for estimating a healthy reference.

Our experiments in out-of-domain settings demonstrate that SCSR \emph{generalizes} effectively without requiring target data adaptation. %
This indicates that SCSR captures robust within-cortex relationships that remain consistent across different datasets. 
Furthermore, SCSR inherently accounts for scanning and processing-related biases within the thickness values, bypassing the need to explicitly model potential confounders \cite{alfaro2021confound}, \revision{as confirmed by our traveling heads experiment and results from harmonization analysis.} 
\revision{
This inherent robustness is reminiscent of brain-asymmetry indices \cite{Kong2018,wachinger2016whole,richards2020increased} that use a subject’s own contralateral hemisphere as an internal reference to achieve invariance.}
\revision{Our results further demonstrate that SCSR is not limited to FreeSurfer-derived thickness maps but can also operate effectively with maps generated by Vox2Cortex. In addition to being faster, Vox2Cortex maps do not require a registration step, which further streamlines the application of SCSR for individualized clinical diagnosis.}
Furthermore, SCSR operates on high-dimensional, \emph{vertex-wise} cortical thickness measurements, allowing for fine-grained localization of atypical patterns.
In contrast, aggregating cortical data into parcels can obscure focal thickness deviations. 
The neural networks in SCSR are multivariate predictors, learning spatial correlations across the cortex and benefiting from the detailed information.
On the other hand, normative models take a mass-univariate approach, estimating separate regression models for each brain measurement, thereby ignoring spatial correlations. 
We implemented SCSR with three different architectures, where the MLP not only delivers competitive accuracy but also offers practical advantages, being easier to train compared to more complex architectures like spherical U-Nets or transformers.
To our knowledge, there is no prior work that utilizes MLPs for predicting vertex-level thickness, highlighting its potential as a viable and efficient alternative for cortical thickness analysis.

Statistical tests in our AD experiments revealed significant differences in Z-scores, particularly within the temporal, parietal, and frontal lobes. Previous studies reported the most significant cortical thinning in AD in the precuneus, entorhinal, parahippocampal, fusiform, inferior temporal, middle temporal, inferior parietal, and frontal regions~\cite{du2007different,schwarz2016large}, all of which are also highlighted by SCSR ($p<0.05$ after Bonferroni correction) in the CN/AD group comparison (cf.~Fig. \ref{fig:AD-summary}). Discriminating MCI patients, who presented intermediate atrophy patterns, remains particularly challenging, but SCSR identified highly significant differences. In particular, we found the medial temporal lobe and regions of the parietal and frontal lobes to be significantly thinner in MCI patients compared to the CN group. A similar result for MCI was previously reported in \cite{Singh2006patternsofcorticalthinning}.  %
Interestingly, the average age of the UKB training data was about 8 years younger than that of the AD datasets, demonstrating the generalization of SCSR to different population distributions. 

Several studies have investigated the automated classification of AD with neural networks \cite{wen2020convolutional}, but these approaches rely on the presence of all diagnoses in the training set, which is problematic because dementia frequently involves co-pathologies, and mixed dementias may display features of multiple conditions \cite{chouliaras2023use}. 
In addition, classification networks function as black-box models, limiting interpretability. 
In contrast, SCSR is not designed for automated classification but for supporting physicians by mapping aberrant cortical thickness. 
SCSR achieves this without the need for disease-specific training, offering a more \emph{transparent} and \emph{disease-agnostic} method suitable for clinical practice. 
Furthermore, the consistency of deviation maps in a longitudinal setting makes them well-suited for tracking disease progression (cf. Fig. \ref{fig:ADlongitudional}). 
In a clinical cohort with dementia patients, SCSR successfully identified atrophy patterns that align with known pathology of dementia types \cite{risacher2013neuroimaging}, not only at the group level but also on an \emph{individual} basis. 
Notably, these results were achieved without any dataset-specific tuning of SCSR.

In our experiments on preterm infants, SCSR identified a significantly lower thickness of the middle temporal gyrus, consistent with prior findings \cite{kelly2024cortical}. 
GAMLSS from BrainCharts, quantile regression, and BLR from PCNtoolkit have not detected significantly lower differences. 
As preterm birth confounds age, it remains unclear whether postmenstrual age or chronological age would be appropriate to map to reference values. 
Unlike these age-dependent models, SCSR bypasses the need for age as a predictor, making it more effective at distinguishing between term and preterm infants, as further evidenced by the t-SNE embeddings. 
In cases with limited training data, such as the dHCP dataset encompassing only about 250 term-born subjects used for training, simpler, parcel-based models with XGBoost can offer a practical balance between accuracy and computational efficiency compared to more complex neural networks.

\emph{Limitations: } %
\revision{While we focused on cortical thickness, other surface-derived features like area, curvature, depth, and myelin mapping could provide additional insights. These metrics could either replace thickness in SCSR, or be concatenated to a multi-channel input.} 
Moreover, we used T1-weighted scans for cortical surface reconstruction in adult experiments, but incorporating T2-weighted scans may further enhance accuracy. Another limitation is that SCSR is not designed to generate brain charts \cite{bethlehem2022brain} but to detect atypical cortical thickness in individuals.
While we demonstrated SCSR's effectiveness for both adult and infant populations, there remains a gap for children and adolescents. 
\revision{Preliminary results indicate that a UKB-trained SCSR does not generalize well to 10-year-old children, requiring further studies on extending SCSR to this age range.} 
Finally, sampling biases common in research studies, including the UK Biobank \cite{fry2017comparison}, need to be kept in mind as they may impact results. 
While our results indicate that SCSR generalizes across ethnicities (trained on UKB, tested on Japanese ADNI), this needs further consideration in the future.

\section{Conclusion}
In conclusion, SCSR addressed key limitations of existing reference models by leveraging deep learning to reconstruct vertex-level cortical thickness maps without relying on demographic data. 
It identified patterns of aberrant cortical thickness across diverse datasets, including dementia patients and preterm infants, where it identified subtle deviations that other models missed.  
On a clinical dataset with four types of dementia, the high spatial resolution enabled the detailed identification of atrophy patterns needed for differential diagnosis. 
Its ability to generate precise, individualized cortical maps from clinical data highlighted its potential for supporting diagnosis in clinical practice.

\section*{Data availability}
Data from the Alzheimer's Disease Neuroimaging Initiative (ADNI) are accessible through the ADNI database (\url{https://adni.loni.usc.edu}) upon registration and compliance with their data usage agreement. 
Data from the Australian Imaging, Biomarkers and Lifestyle (AIBL) study are available upon request at \url{https://aibl.org.au/}.
Data from the Japanese Alzheimer's Disease Neuroimaging Initiative (J-ADNI) can be obtained upon registration and adherence to their data usage policies through their official website (\url{https://www.j-adni.org/}).
 Data from the German Longitudinal Cognitive Impairment and Dementia Study (DELCODE) are accessible by request through the German Center for Neurodegenerative Diseases (DZNE) website (\url{https://www.dzne.de/en/research/studies/clinical-studies/delcode/}), following their data access procedures.
  The Developing Human Connectome Project (dHCP) data are available for download from their website (https://data.developingconnectome.org/) upon registration and agreement to their data usage terms.
UK Biobank (UKBB) data can be requested via the UKBB website (\url{https://www.ukbiobank.ac.uk/}) following their application process. 
The in-house dataset can be shared upon request.
Trained models are available at the GitHub repository. 

\section*{Code availability}
Source code is available at \url{https://github.com/ai-med/SCSR-core}.

\section*{Acknowledgment}
This research was partially supported by the German Research Foundation and the DAAD programme Konrad Zuse Schools of Excellence in Artificial Intelligence, sponsored by the Federal Ministry of Research, Technology and Space.
The authors gratefully acknowledge the Leibniz Supercomputing Centre by providing computing time on its Linux Cluster.

We thank Jakob Seidlitz, Aaron Alexander-Bloch and Lena Dorfschmidt (Department of Child and Adolescent Psychiatry and Behavioral Science, The Children's Hospital of Philadelphia, Philadelphia, PA, 19104 USA) for sharing and providing help with BrainChart models, and Chris Adamson and Gareth Ball (Murdoch Children’s Research Institute, Melbourne, Victoria, Australia) for providing the MCRIBS parcellation of the dHCP dataset. 

Data collection and sharing for this project was funded by the Alzheimer's Disease
Neuroimaging Initiative (ADNI) (National Institutes of Health Grant U01 AG024904) and
DOD ADNI (Department of Defense award number W81XWH-12-2-0012). ADNI is funded
by the National Institute on Aging, the National Institute of Biomedical Imaging and
Bioengineering, and through generous contributions from the following: Alzheimer's
Association; Alzheimer's Drug Discovery Foundation; Araclon Biotech; BioClinica, Inc.;
Biogen Idec Inc.; Bristol-Myers Squibb Company; Eisai Inc.; Elan Pharmaceuticals, Inc.; Eli
Lilly and Company; EuroImmun; F. Hoffmann-La Roche Ltd and its affiliated company
Genentech, Inc.; Fujirebio; GE Healthcare; ; IXICO Ltd.; Janssen Alzheimer Immunotherapy
Research \& Development, LLC.; Johnson \& Johnson Pharmaceutical Research \&
Development LLC.; Medpace, Inc.; Merck \& Co., Inc.; Meso Scale Diagnostics,
LLC.; NeuroRx Research; Neurotrack Technologies; Novartis Pharmaceuticals
Corporation; Pfizer Inc.; Piramal Imaging; Servier; Synarc Inc.; and Takeda Pharmaceutical
Company. The Canadian Institutes of Health Research is providing funds to support ADNI
clinical sites in Canada. Private sector contributions are facilitated by the Foundation for the
National Institutes of Health (www.fnih.org). The grantee organization is the Northern
California Institute for Research and Education, and the study is coordinated by the
Alzheimer's Disease Cooperative Study at the University of California, San Diego. ADNI
data are disseminated by the Laboratory for Neuro Imaging at the University of Southern
California.

J-ADNI was supported by the following grants: Translational Research Promotion Project from the New Energy and Industrial Technology Development Organization of Japan; Research on Dementia, Health Labor Sciences Research Grant; Life Science Database Integration Project of Japan Science and Technology Agency; Research Association of Biotechnology (contributed by Astellas Pharma Inc., Bristol-Myers Squibb, Daiichi-Sankyo, Eisai, Eli Lilly and Co., Merck-Banyu, Mitsubishi Tanabe Pharma, Pfizer Inc., Shionogi \& Co., Ltd., Sumitomo Dainippon, and Takeda Pharmaceutical Company), Japan, and a grant from an anonymous foundation.

Data used in the preparation of this article were obtained from the Australian Imaging Biomarkers and Lifestyle Flagship Study of Ageing (AIBL) funded by the Commonwealth Scientific and Industrial Research Organisation (CSIRO), which was made available at the ADNI database (www.loni.ucla.edu/ADNI).

Data used in the preparation of this article were obtained from the UK Biobank Resource under Application No. 34479.

\bibliographystyle{elsarticle-harv}
\bibliography{refs}

\clearpage
\appendix
\counterwithin{figure}{section}
\counterwithin{table}{section}
\renewcommand{\thetable}{A.\arabic{table}}
\renewcommand{\thefigure}{A.\arabic{figure}}

\newgeometry{left=2cm, right=2cm}

\section{Supplementary Material}

\subsection{Dataset details \label{sec:datadetails}}

\noindent
\textbf{UKB:} We restricted our analyses to neurologically healthy individuals by excluding participants with any evidence of brain disease or head cancer. Specifically, subjects were excluded if they had a self-reported or clinically recorded history of any of the following conditions: dementia, multiple sclerosis, stroke, traumatic brain injury, brain abscess, motor neuron disease, epilepsy, depression, anxiety, schizophrenia, bipolar disorder, or Parkinson’s disease. In addition, individuals with a history of cancer, particularly those with head and neck cancers, were also removed from the cohort. 

\noindent
\textbf{ADNI:} 
Data used in preparation of this article were obtained from the Alzheimer's Disease Neuroimaging Initiative (ADNI) database (adni.loni.usc.edu). As such, the investigators within the ADNI contributed to the design and implementation of ADNI and/or provided data but did not participate in analysis or writing of this report.
A complete listing of ADNI investigators can be found at:
\url{http://adni.loni.usc.edu/wp-content/uploads/how_to_apply/ADNI_Acknowledgement_List.pdf}

\noindent
\textbf{AIBL:}
Data used in the preparation of this article was obtained from the Australian Imaging Biomarkers and Lifestyle flagship study of ageing (AIBL) funded by the Commonwealth Scientific and Industrial Research Organisation (CSIRO) which was made available at the ADNI database (www.loni.usc.edu/ADNI). The AIBL researchers contributed data but did not participate in analysis or writing of this report. AIBL researchers are listed at www.aibl.csiro.au.

\noindent
\textbf{DELCODE:}
Data used in the preparation of this article were obtained from the study group DELCODE of Clinical Research from the German Center for Neurodegenerative Diseases (DZNE).

\noindent
\textbf{J-ADNI:}
Data used in the preparation of this article were obtained from the Japanese Alzheimer’s Disease Neuroimaging Initiative (J-ADNI)database deposited in the National BioscienceDatabase Center Human Database, Japan(Research ID: hum0043.v1, 2016). As such, the investigators within J-ADNI contributed to the design and implementation of J-ADNI or provided data but did not participate in the analysis or writing of this report. A complete listing of J-ADNI investigators can be found at  \url{https://humandbs.dbcls.jp/en/hum0043-v1}.

\begin{figure}[b]
    \centering
    \includegraphics[width=0.5\textwidth]{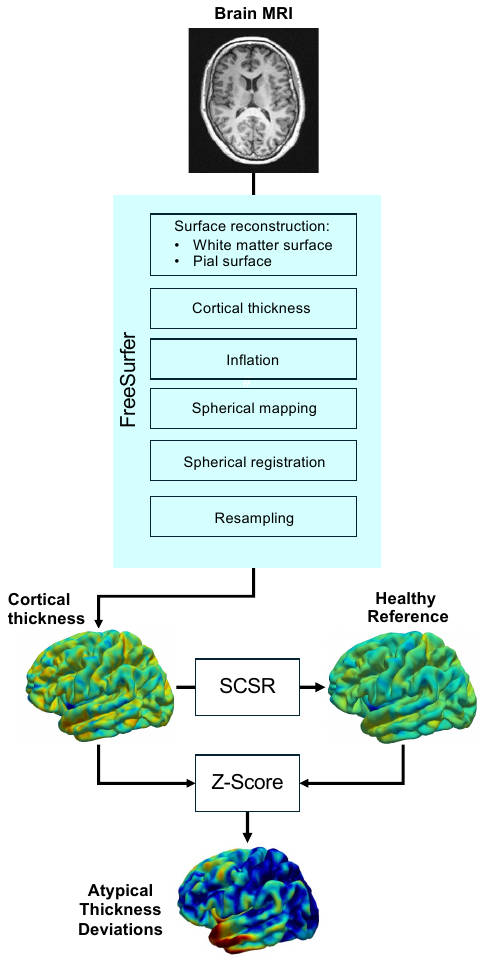}
    \caption{Processing pipeline for detecting atypical thickness deviations starting from a brain MRI scan. The surface-based processing stream in FreeSurfer \cite{fischl2012freesurfer} is used to extract cortical thickness maps. The registration to the template fsverage6 is part of it, creating correspondences across subjects. The FreeSurfer output serves as input to SCSR for estimating the healthy reference and, finally, for the Z-score computation. }
    \label{fig:freesurfer}
\end{figure}

\setlength{\tabcolsep}{5pt}
\begin{table}
    \centering
    \rowcolors{2}{gray!15}{white}
    \begin{tabular}{lrccrrrrrrr}
    \toprule
    Dataset    & Subjects & Sex (\% M/F) & Age (Mean, SD) & CN    & MCI   & AD    & bvFTD & PCA   & SD    \\
    \midrule
    UK Biobank & 31,673   & 47.3 / 52.7  & 64.0 (7.5)     & 31,673 & 0     & 0     & 0     & 0     & 0     \\
    ADNI       & 1,226    & 53.5 / 46.5  & 73.0 (7.2)     & 343    & 665   & 218   & 0     & 0     & 0     \\
    AIBL       & 216      & 39.4 / 60.6  & 70.8 (6.2)     & 157    & 42    & 17    & 0     & 0     & 0     \\
    DELCODE    & 386      & 46.4 / 53.6  & 71.3 (5.9)     & 174    & 141   & 71    & 0     & 0     & 0     \\
    J-ADNI     & 428      & 47.0 / 53.0  & 71.9 (6.4)     & 114    & 208   & 106   & 0     & 0     & 0     \\
    In-House     & 50      & 34.0 / 66.0  & 63.1 (8.0)     & 10    & 0   & 10   & 10     & 10     & 10     \\
    \bottomrule
    \end{tabular}    
    \caption{Subject characteristics of the datasets.%
    }
    \label{tab:datasets}
\end{table}

\begin{figure}[b]
    \centering
    \includegraphics[width=0.99\textwidth]{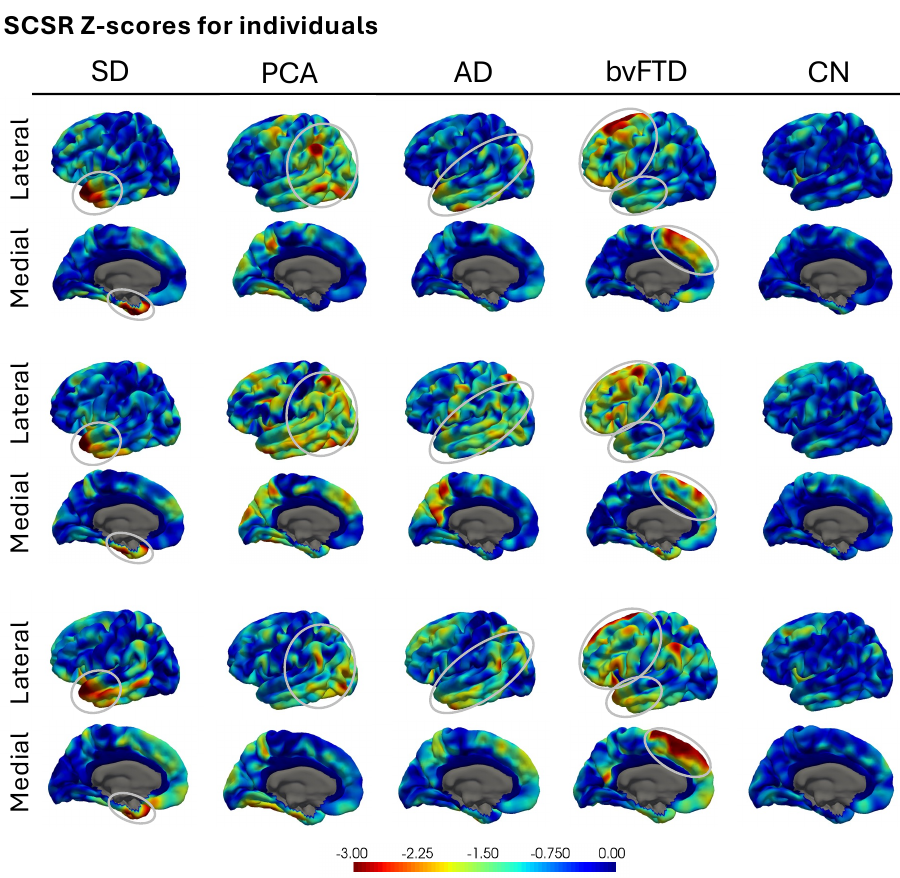}
    \caption{\revision{Extension of Fig. \ref{fig:diff-diag} by adding medial views to the previously shown lateral views.}  Differential diagnosis of dementia on routine clinical data, covering Alzheimer's disease (AD), posterior cortical atrophy (PCA), behavioral variant frontotemporal dementia (bvFTD), semantic dementia (SD), and cognitively normal (CN). Lateral and medial views of individual Z-scores across diagnostic groups. Ellipses highlight regions with the strongest atrophy.}
    \label{fig:diffDiag_supp}
\end{figure}

\begin{figure}[b]
    \centering
    \includegraphics[width=0.99\textwidth]{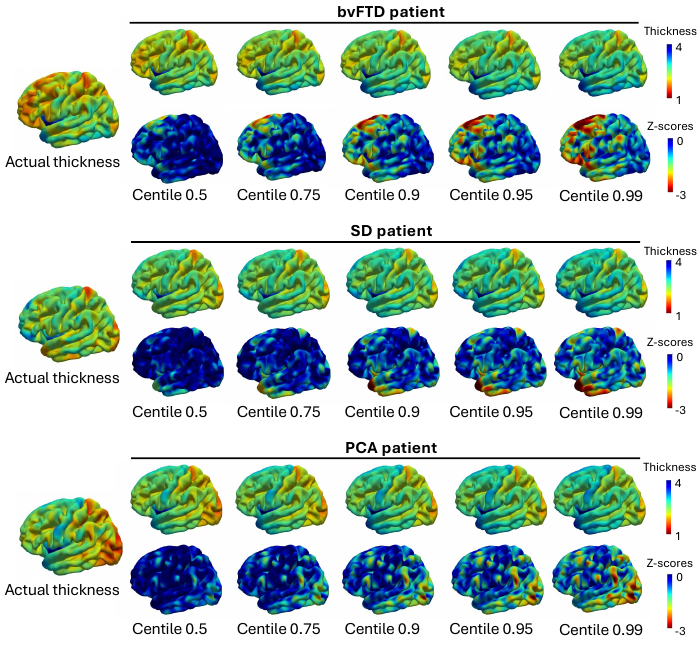}
    \caption{Cortical analysis for three patients diagnosed with bvFTD, SD, and PCA, respectively. 
    We show the actual cortical thickness surface, SCSR reference reconstructions, and Z-score maps across reconstruction centiles from 0.5 to 0.99. 
The SCSR references show significant inter-subject differences, reinforcing our findings from the reconstruction error that subject-specific references are created. As the reconstruction centiles increase, a progressive increase in cortical thickness is observed in specific regions, accompanied by more distinct atrophy patterns in the Z-score maps.   }
    \label{fig:recons}
\end{figure}

\begin{table}[!ht]
    \setlength{\tabcolsep}{2pt}
    \centering
    \caption{Comparison of SCSR architectures with two neural networks (MLP, spherical U-Net) and XGBoost on the AD test set. Experimental setting is comparable to Fig. \ref{fig:AD-ROI}. \label{tab:architectures}}    
    \begin{tabular}{lcccccccccc}
    
    \toprule
    
    & \multicolumn{5}{c}{ROC AUC} & \multicolumn{5}{c}{Spearman correlation} \\ 
    \cmidrule(lr){2-6}
    \cmidrule(lr){7-11}

    ~ & All & ADNI  & AIBL & DELCODE & JADNI & All & ADNI  & AIBL & DELCODE & JADNI  \\ 
    \midrule
        SCSR MLP & 0.80 & 0.77 & 0.74 & 0.80 & 0.81 & -0.54 & -0.42 & -0.34 & -0.57 & -0.59  \\ 
        SCSR SU-Net & 0.79 & 0.76 & 0.75 & 0.80 & 0.81 & -0.53 & -0.39 & -0.39 & -0.60 & -0.57  \\ 
        SCSR XGBoost & 0.77 & 0.76 & 0.76 & 0.80 & 0.77 & -0.46 & -0.40 & -0.35 & -0.58 & -0.51 \\ 

    \bottomrule
    \end{tabular}
\end{table}

\begin{figure}[b]
    \centering
    \includegraphics[width=0.99\textwidth]{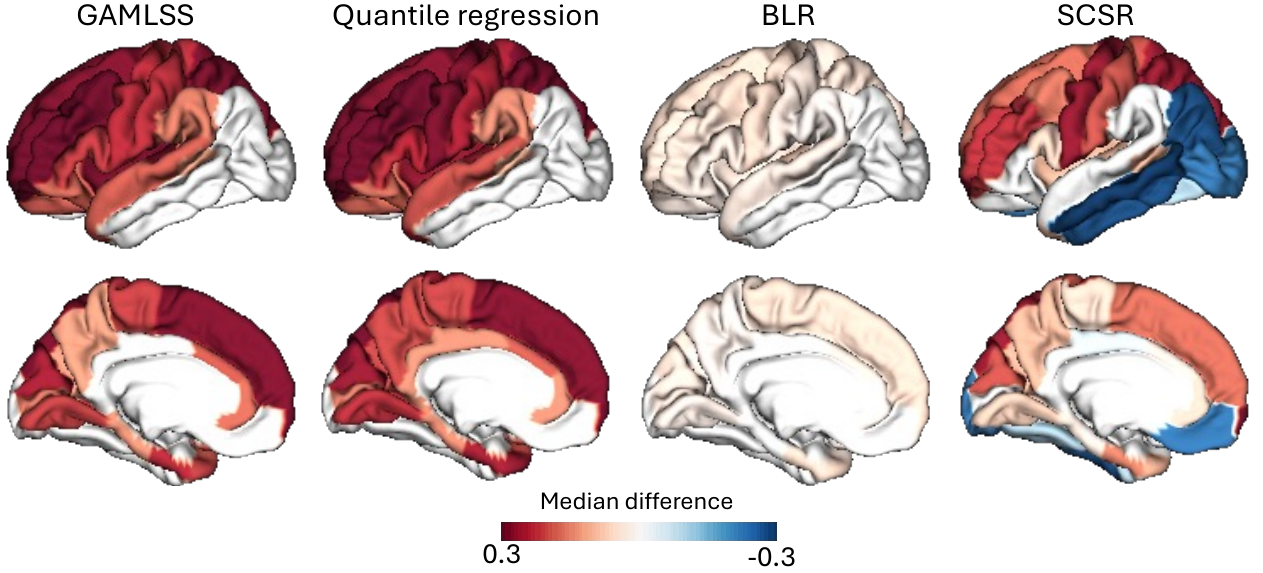}
    \caption{Lateral and medial views of group comparison results from Fig. \ref{fig:preterm}. Cortical renderings were generated with the ENIGMA toolbox in Python \cite{lariviere2021enigma}. }
    \label{fig:preterm_full}
\end{figure}

\begin{figure}
    \centering
    \includegraphics[width=0.75\textwidth]{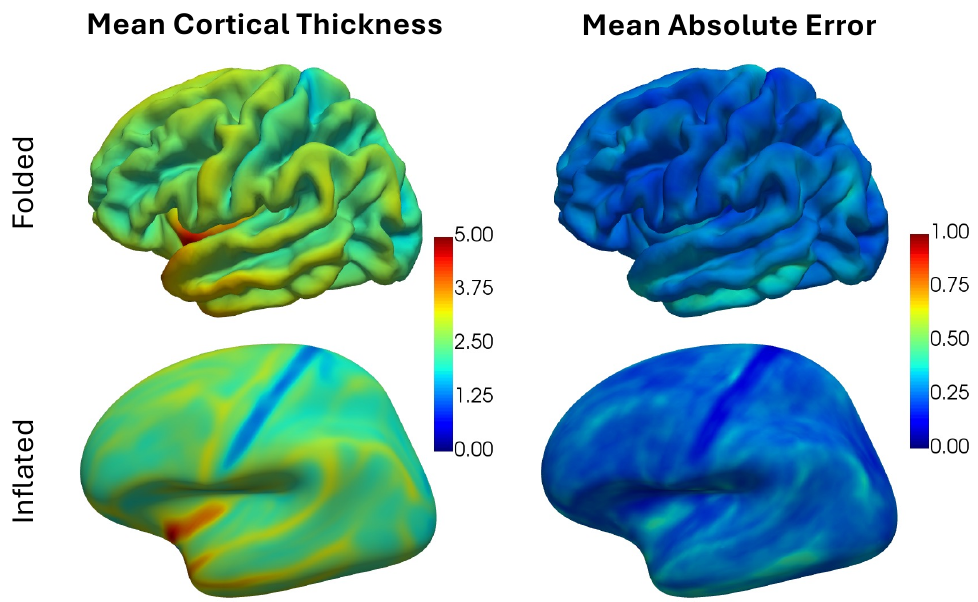}
    \caption{\revision{Mean ground-truth cortical thickness and mean reconstruction error from SCSR (sampling rate 20\%, $q=0.95$) across 1,000 randomly chosen UKB training scans. The same vertex-wise maps are shown on the folded and the inflated FsAverage template. Values are in mm.}}
    \label{fig:ukb_train_error}
\end{figure}

\end{document}